\documentclass[10pt,conference]{IEEEtran}
\IEEEoverridecommandlockouts
\usepackage{cite}
\usepackage{amsmath,amssymb,amsfonts}

\usepackage[noend]{algorithm,algpseudocode}

\usepackage{graphicx}
\usepackage{textcomp}
\usepackage{xcolor}
\usepackage{subfig}

\algnewcommand\algorithmicforeach{\textbf{for each}}
\algdef{S}[FOR]{ForEach}[1]{\algorithmicforeach\ #1\ \algorithmicdo}

\def\BibTeX{{\rm B\kern-.05em{\sc i\kern-.025em b}\kern-.08em
    T\kern-.1667em\lower.7ex\hbox{E}\kern-.125emX}}

\IEEEoverridecommandlockouts\IEEEpubid{\makebox[\columnwidth]{ 978-1-6654-3540-6/22/\$31.00~\copyright~2022 IEEE \hfill} \hspace{\columnsep}\makebox[\columnwidth]{ }}    
    
\begin{document}

\title{An Explainer for Temporal Graph Neural Networks}
%\thanks{}
%}

\author{\IEEEauthorblockN{1\textsuperscript{st} Wenchong He*}
\IEEEauthorblockA{\textit{University of Florida}\\
Gainesville, USA. \\
whe2@ufl.edu}
\and
\IEEEauthorblockN{1\textsuperscript{st} Minh N. Vu*}\thanks{*Authors contribute equally}
\IEEEauthorblockA{\textit{University of Florida}\\
Gainesville, USA.\\
minhvu@ufl.edu}
\and
\IEEEauthorblockN{3\textsuperscript{rd} Zhe Jiang}
\IEEEauthorblockA{\textit{University of Florida} \\
Gainesville, USA. \\
zhe.jiang@ufl.edu}
\and
\IEEEauthorblockN{4\textsuperscript{th} My T. Thai}
\IEEEauthorblockA{\textit{University of Florida} \\
Gainesville, USA. \\
mythai@cise.ufl.edu}
}

\maketitle

\begin{abstract}
Temporal graph neural networks (TGNNs) have been widely used for modeling time-evolving graph-related tasks due to their ability to capture both graph topology dependency and non-linear temporal dynamic. The explanation of TGNNs is of vital importance for a transparent and trustworthy model. However, the complex topology structure and temporal dependency make explaining  TGNN models very challenging. In this paper, we propose a novel explainer framework for TGNN models. Given a time series on a graph to be explained, the framework can identify dominant explanations in the form of a probabilistic graphical model in a time period. Case studies on the transportation domain demonstrate that the proposed approach can discover dynamic dependency structures in a road network for a time period. 
\end{abstract}


\section{Introduction}

In the last several years, 
Graph Neural Networks (GNNs) have been increasingly popular due to their ability to capture the complex relationship and interactions in a system\cite{Defferrard2016,Kipf2016iclr,GraphSage,velickovic2018graph,CayleyNets,MOTIFNET, xu2018how,Ying_NIPS2018,xinyi2018capsule}. GNNs assume static graph data and operate by aggregating information in the local neighborhood of a node. 
However, many real-world systems are dynamic and evolving over time. %, such as transportation networks and social networks. 
For example, in the transportation domain, accurate traffic forecasting requires both temporal features (periodicity and trend) and spatial dependency modeling (the topology of a road network). A Temporal Graph Neural Network (TGNN) combines both a GNN and a recurrent neural network (RNN) to capture both the spatial  dependency and temporal dynamics in a system\cite{zhao2019t,wang2020traffic,min2021stgsn}. 

Although TGNNs have achieved wide success in modeling spatial networks with temporal dynamics, it is unclear why a model makes certain predictions. Such explanation is crucial because: (1) It improves the transparency and consequently the trust of a TGNN model in a growing number of safety-related applications when being deployed into the real world; (2) It allows end-users to  identify  interesting dynamic dependency structures in the system.  For example, for traffic forecasting, the spatial dependency structure can change due to the evolving characteristics of the traffic flow.  In a non-congestion period, the traffic status at upstream roads impacts the traffic
status at downstream roads through the transfer effect,  and in a congestion period, the traffic status at downstream roads impacts the traffic status
at upstream roads through the feedback effect\cite{dong2012spatial,zhao2019t}. Such dependency dynamics are crucial for a transportation domain expert to identify the influential neighboring roads at different time periods when predicting the traffic flow or speed for one road segment.  

The problem of explaining TGNNs is challenging for several reasons. First, to generate an accurate explanation for the temporal predictions, we need to consider the network structural and temporal dependency simultaneously, which increases the problem complexity compared with the static GNN explanations. Second, the explanations identified in some time steps may be  insignificant or redundant, it is non-trivial to discover the dominant interesting dependency structure in a time period and eliminate the redundant pattern. Third, the computation complexity is  high considering the high complexity of the interpretable domain and quadratic potential temporal windows for explanations.

 We propose a novel explainer framework for the temporal graph neural network interpretations based on the static graph explainer. Specifically, the main contributions are:
 
$\bullet$ To reduce the interpretation complexity, we reformulate the temporal graph neural network model such that the explanations at different temporal snapshots can be done independently to leverage the static graph explainer. 

$\bullet$ We propose a pruning approach to discover the temporal dominant interesting explanations from the independent explanations at each time step and eliminate the insignificant and redundant explanations. The proposed pruning algorithms can reduce the temporal search space and improve efficiency. 

$\bullet$ The  case studies of the  TGNN explanations on the transportation domain show that our proposed explainer framework can provide interpretable explanations and discover the dynamic dependency structure of the spatial network.

% Problem definition: 

% Motivation: traffic road network; brain disease

% Challenge: 

% Related work

\section{Background on Graph Neural Network Explanation}
This section provides some related background and preliminaries of our paper. The following provides some highlights on GNN and TGNN, the current landscape of explaining predictions made by GNNs, and the PGM-Explainer, which serves as an important component of our method.

\textbf{GNNs and TGNNs:} With the increasing availability of modern graph data and the popularity of graph-related tasks, many Graph-based Neural Networks have been introduced in recent years, such as ChebNets \cite{Defferrard2016}, Graph Convolutional Networks \cite{Kipf2016iclr}, GraphSage \cite{GraphSage}, Graph Attention Networks \cite{velickovic2018graph}, among others~\cite{CayleyNets,MOTIFNET, xu2018how,Ying_NIPS2018,xinyi2018capsule, Lee2019}. At a high level, these models exploit the non-linear transformation in conventional neural networks to transform input features and aggregate (or propagate) those features via a graph's connectivity. To consider the time-varying features of graph-related tasks, TGNNs combine GNN and RNN models together, e.g., Temporal Graph Convolutional Neural Network (T-GCN)\cite{zhao2019t}, Spatial-Temporal Graph Social Network (STGSN)\cite{min2021stgsn}. A TGNN  integrates the GNN architecture (to model the structural topology dependency) and an RNN model (to capture the non-linear temporal dependency for time series).  Recent studies have put significant attention into the TGNN model, such as applications in traffic forecasting, social network analysis, and human trajectory prediction \cite{zhao2019t,min2021stgsn,mohamed2020social}. 

% \vnm{to Wenchong: Can you add a paragraph/a few sentences here briefly describe the TGNN? We need something to tell reader what is the key differences of TGNN compared to static ones.}

\textbf{Explanation methods for GNNs:} Similar to some previously studied deep neural network architectures, the GNNs are known to be \textit{black-boxes}%~\cite{Yuan2020ExplainabilityIG, Ying_NIPS2019}
, i.e. it is unclear how they generate predictions. To address this issue, several explanation methods, called \textit{explainers}, for GNNs have been introduced recently~\cite{Ying_NIPS2019, minhpgm, Schnake2021}. The main objective of the explainers is to identify some network components such as nodes, edges, or sub-graphs, that explain, clarify or contribute the most to the model's predictions. The scope of explanations can also vary: some methods focus on explaining the overall behaviors of the model, while others aim to explain the model's prediction on a specific input instance. In this work, we focus on the latter, which is known as the ``local explanation" method. Local explanation methods can also be classified based on their methodology: gradients or features-based,  perturbation-based, decomposition, and surrogate methods. %A more comprehensive review can be found in a recent survey~\cite{Yuan2020ExplainabilityIG}.   

% However, all existing methods are not designed for temporal models.

% \textbf{Probabilistic Graphical Explanations of Graph-based Neural Network:} 

% There are many non-temporal Graph-based Neural Networks (GNNs) have been introduced and deployed in recent years such as ChebNets \cite{Defferrard2016}, Graph Convolutional Networks \cite{Kipf2016iclr}, GraphSage \cite{GraphSage}, Graph Attention Networks \cite{velickovic2018graph}, among others~\cite{CayleyNets,MOTIFNET, xu2018how,Ying_NIPS2018,xinyi2018capsule, Lee2019}. In general, they can be described by a more specific form of (\ref{eq:tgcn}): $  Y = \Phi(G; X)$. The problem of explaining GNNs, i.e. the function $\Phi$,  has been addressed by several different explanation methods, called explainers~\cite{Ying_NIPS2019, minhpgm, Schnake2021}. The main objectives of these methods is identifying some network's components, i.e. nodes, edges or sub-graphs, that contributed mostly to the model's predictions. However, all existing methods are not designed for temporal models. 

\textbf{Explaining the GNN using PGM-Explainer:} This paper aims to extend PGM-Explainer~\cite{minhpgm}, a local explanation method based on a probabilistic graphical model,  from GNN explanation to TGNN explanation. The main reason why we choose PGM-Explainer as a base framework is that it can capture the graph dependencies within TGNN's variables. This makes the explainer a strong candidate to explain models with high temporal-spatial dependencies as TGNNs. In the next paragraphs, we provide a more formal description of how PGM-Explainer explains the static GNNs. 

 We consider the forwarding of a GNN or TGNN to be explained as a function $\Phi$. In practice, since the output of $\Phi$ can contain many predictions (for example as in node classification), we denote $o$ the target prediction to be explained. As such, the prediction to be explained can be written as the function $\Phi_o: \mathcal{G}\rightarrow \mathcal{K}$, where $ \mathcal{G}$ is the set of input graphs and $\mathcal{K}$ is the set of prediction's outputs on that target.

 PGM-Explainer represents the GNN's variables via a directed acyclic probabilistic graph, i.e. a Bayesian network~\cite{PEARL198877}. Given a target prediction to be explained $o$, PGM-Explainer solves the optimal Bayesian network $\mathcal{B}^*$:
\begin{align}
    &\arg \max_{\mathcal{B} \in \mathcal{E}} R_{\Phi, o}(\mathcal{B}), \quad
    \textup{s.t. }  |\mathcal{V}(\mathcal{B})| \leq M, \boldsymbol{o} \in \mathcal{V}(\mathcal{B}), \label{ex_optimization}
\end{align}
where $\mathcal{E}$ is the set of all Bayesian networks and $\boldsymbol{o}$ is the random variable corresponding to the target prediction $o$. The set of random variables in Bayesian network $\mathcal{V}(\mathcal{B})$ represents the set of explanatory features in the input graph of the model, which is typically a subset of nodes in the input graph. The objective $R_{\Phi, o}(\mathcal{B})$ measure the fitness of the Bayesian network $\mathcal{B}$ with a set of perturbation data generated from the function $\Phi_o$. The constraints in the optimization is to encourage compact solutions and guarantee the target prediction is included in the explanations. One choice of the objective $R_{\Phi, o}(\mathcal{B})$ is the \textit{BIC score}, which is given as follows:
\begin{align*}
     R_{\Phi, o}(\mathcal{B}) = \textup{score}_{BIC} (\mathcal{B}  :  \mathcal{D}_o) =  l(\hat{{\theta}}_{\mathcal{B}}  :  \mathcal{D}_o) - \frac{\textup{log} n}{2}\textup{Dim}[\mathcal{B}] 
\end{align*}
where $\mathcal{D}_o$ is the perturbation data generated from $\Phi_o$ and $\textup{Dim}[\mathcal{B}]$ is the dimension of model $\mathcal{B}$. ${\theta}_{\mathcal{B}}$ are the parameters of
$\mathcal{B}$ and function $l({\theta}_{\mathcal{B}}  :  \mathcal{D}_o)$ is the log-likelihood between the data $\mathcal{D}_o$ and  ${\theta}_{\mathcal{B}}$. $\hat{{\theta}}_{\mathcal{B}}$ in the score is parameters' value that maximizes the log-likelihood, which is called the maximum likelihood estimator. This choice of objective has been proven to be \textit{consistent} with the data~\cite{Koller2009}.

% In fact, PGM-Explainer represents the model's variables via a directed acyclic probabilistic graph, i.e. the Bayesian network~\cite{PEARL198877}. Given target prediction to be explained $o$, PGM-Explainer solves for the optimal Bayesian network $\mathcal{B}^*$:
% \begin{align}
%     &\arg \max_{\mathcal{B} \in \mathcal{E}} R_{\Phi, o}(\mathcal{B}), \quad
%     \textup{s.t. }  |\mathcal{V}(\mathcal{B})| \leq M, \boldsymbol{o} \in \mathcal{V}(\mathcal{B}), \label{ex_optimization}
% \end{align}
% where $\mathcal{E}$ is the set of all Bayesian networks and $\boldsymbol{o}$ is the random variable corresponding to the target prediction $o$. The set of random variables in Bayesian network $\mathcal{V}(\mathcal{B})$ represents the set of explanatory features in the input graph of the model, which is typically a subset of nodes in the input graph. The constraints in the optimization is to encourage compact solutions and guarantee the target prediction is included in the explanations. 

\section{Problem Statement}  \label{sect:statement}
This section provides a problem formulation for TGNN explanation. Based on our previous introduction of PGM-Explainer in case of static GNNs,  we describe the formulation for TGNN and highlight some key properties and challenges of the problem compared to the static case.

% To leverage PGM-Explainer to TGNN, we need to (1) given a T-GCN and the target of explained $o$, specify the function $\Phi_o$, (2) describe how the perturbation data $\mathbf{D}$ can be generated from $\Phi_o$ and (3) specify the objective $R_{\Phi, o}$.

 The function learnt by a TGNN can be considered as a mapping from a graph $G$ and a sequence of feature matrices to the predictions:
\begin{align}
    Y = \Phi (G; [X_{t}, \cdots, X_{t+T-1}]), \label{original}
\end{align}
where  $T$ is the time window of the model. In practice, the total time interval $[t_1,t_2]$ is normally larger than the window $T$. In that case, the interval $[t_1,t_2]$ is processed into $t_2 - t_1 - T + 1$ sequences of length $T$ and feed into the TGNN. We consequentially have $t_2 - t_1 - T + 1$ predictions. 

Therefore, the problem of explaining the TGNN's predictions is not only about explaining the function $\Phi$ at some specific input sequence $[X_{t}, \cdots, X_{t+T-1}]$ but also about explaining all predictions during the time interval $[t_1,t_2]$, which includes $t_2 - t_1 - T + 1$ predictions. 

Formally, given a trained TGNN model $\Phi$ and the predictions at the temporal interval to be explained,  
we define our problems as discovering the crucial dependency structures for the model making predictions on the interval. In the followings, we will first address how prediction of each sequence of length $T$ is explained. Then, we describe how all explanations can be combined to explain all predictions on the interval.

\section{Approach}
\begin{figure*}
    \centering
    {\includegraphics[height=1.8in]{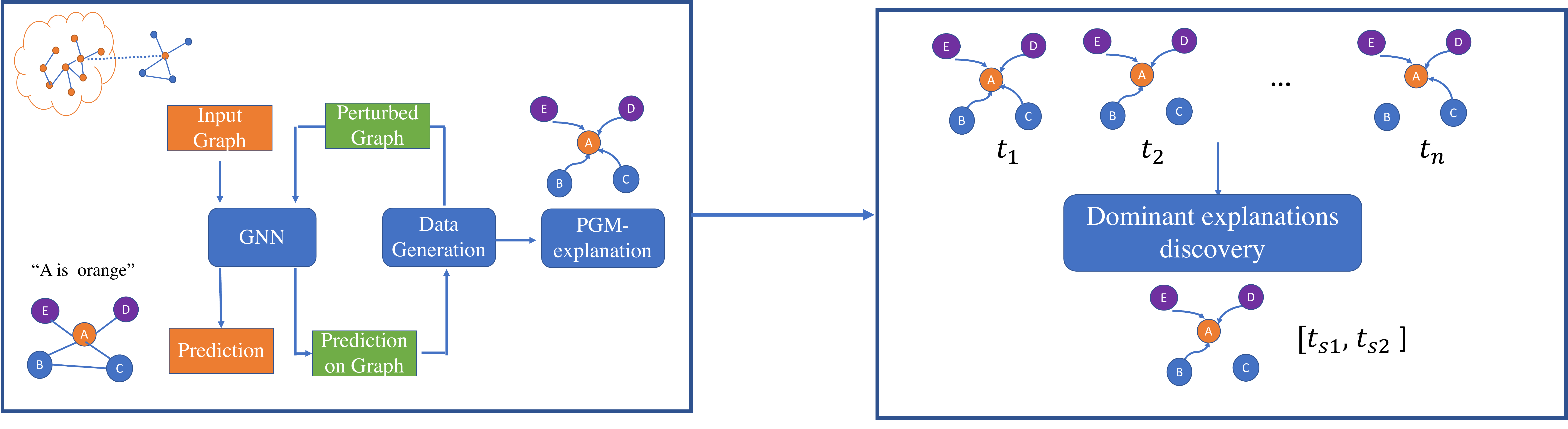}} 
    \caption{Our proposed temporal graph neural network explainer framework }
    \label{fig:framework}
\end{figure*}

%We propose a novel explainer framework for explaining TGNN model. 
Our framework consists of two modules as shown in Figure~\ref{fig:framework}. The first module leverages PGM-explainer to explain TGNN model independently for each time step. Then the second module aims to discovery the dominant interesting explanations from the explanations identified in the first module. In the following we will discuss the two modules and use T-GCN \cite{zhao2019t} as an example of TGNN model. 

\subsection{T-GCN Explanation}
%%% Minh finish this part
A naive way to explain the TGNN $\Phi$ on its input $[X_{t}, \cdots, X_{t+T-1}]$ is to treat the sequence as a single feature matrix and explain the model as in the case of non-temporal models. One drawback of this approach is the temporal information and dependencies among model's variables are not considered and included in the explanations. Furthermore, the number of input's features would be scaled up with a factor of the window $T$. This introduces significant computation complexity on the explaining process. 

To overcome challenges in explaining the TGNN discussed in Sect.~\ref{sect:statement} , we reformulate the TGNN equations based on its sequential implementation~\cite{zhao2019t} as follow:
\begin{align}
    Y &= \bar{\Phi} (G; X_{t+T-1}; H_{t+T-1}), \label{eq:tgnnlast}\\
    H_{i+1} &= \bar{\Phi} (G; X_{i}; H_{i}), i=t,\cdots ,t + T -2. \label{outeq}
\end{align}
Here, $H_i$ is the hidden features or configurations that the TGNN computes during the sequential forwarding computation. With this formulation, we can consider the problem of explaining the TGNN $\Phi$ as the problem of explaining $\bar{\Phi}$ at its last computation with argument $H_{t+T-1}$. On one hand, the argument $H_{t+T-1}$ is the result of the aggregation of the input features from the past time steps. On the other hand, it can be considered as the configuration of the last computation where the model makes the final prediction. That configuration dictates how the graph's components work together to generate the model's prediction. Thus, we rewrite (Eq.~\ref{eq:tgnnlast}) as $Y =  \bar{\Phi}_{ H_{t+T-1}} (G; X_{t+T-1})$. Then, for each sequence $[X_{t}, \cdots, X_{t+T-1}]$ the function $\bar{\Phi}_{ H_{t+T-1}}$ can be explained by PGM-Explainer as described in Eq.~\ref{ex_optimization}. 

Specifically, for each time sequence starting with index $t$, we generate a perturbation data $\mathcal{D}_t$ of $\bar{\Phi}_{H_{t+T-1}}$. Similar to that in PGM-Explainer, $\mathcal{D}_t$ contains two components. First is a set of random seeds determining which nodes of the input graph $G$ is perturbed. Second is an indicator whether the predictions on each nodes is changed by the perturbation on the nodes. The predictions on the perturbations are computed based on the function $\bar{\Phi}_{H_{t+T-1}}$, in which the value of $H_{t+T-1}$ is computed based on the original input data by forwarding the model using Eq.~\ref{outeq}. For each $\mathcal{D}_t$, we use PGM-Explainer to solve for an optimal Bayesian network $\mathcal{B}_t$ explaining the model at that snapshot. 

% Equations~\ref{original} and \ref{eq:tgnnlast} describe the TGNN prediction of one time sequence. In practice, the TGNN is normally used to compute predictions in some given time intervals larger than the window $T$. We denote $t_s$ and $t_e$ the starting indices of the first and the last sequence of length $T$ in the interval. As such, for each input sequence, we have a perturbation $\mathcal{D}_{t}$ and a Bayesian network explanation $\mathcal{B}_t$, $t = t_s,\cdots, t_e$.

Note that the explanation Bayesian network $\mathcal{B}_t$ can be different among different snapshots. In analyzing temporal predictions of TGNN, the network structures that can represent many snapshots are much more favorable. Thus, we propose a Temporal Bayesian Information Criterion (TBIC) score measuring the fitness of a given Bayesian network $\mathcal{B}$ with a temporal data $\mathbf{D} =\{\mathcal{D}_{t}\}_{t=t_s}^{t_e}$:
\begin{align*}
\textup{score}_{TBIC} (\mathcal{B}  :  \mathbf{D}) := 
    \frac{1}{t_e - t_s + 1}\sum_{t = t_s}^{t_e}
    \textup{score}_{BIC}(\mathcal{B}: \mathcal{D}_{t}).
\end{align*}
To mitigate the impact of variations in the data $\mathcal{D}_t$ at each snapshot and grasp a better intuition on how much the edges in the Bayesian network $\mathcal{B}$ help capture the data, we use a normalized version of the TBIC score, called $F_{\mathbf{D}}(\mathcal{B})$. The normalized TBIC is the different between the TBIC score of $\mathcal{B}$  and the Bayesian network with no edge $\mathcal{B}_0$:
\begin{align}
    &F_{\mathbf{D}}(\mathcal{B}) :=  \textup{score}_{TBIC} (\mathcal{B}  :  \mathbf{D}) - \textup{score}_{TBIC} (\mathcal{B}_0  :  \mathbf{D}) \label{eq:tbicnorm}
    \\
    = & \frac{1}{t_e - t_s + 1}\sum_{t = t_s}^{t_e} \left(
    \textup{score}_{BIC}(\mathcal{B}: \mathcal{D}_{t}) - \textup{score}_{BIC}(\mathcal{B}_0: \mathcal{D}_{t}) \right). \nonumber
\end{align}
We can see that each term in the sum of Eq.~\ref{eq:tbicnorm} captures the gain of including the edges in $\mathcal{B}_t$ to fit each data $\mathcal{D}_t$.

With the normalized TBIC score $F_{\mathbf{D}}$, we now can define the \textit{interesting Bayesian network} and the \textit{temporal dominant interesting Bayesian network}. Specifically, the Bayesian network $\mathcal{B}$ is an \emph{interesting} Bayesian network on a given temporal window $[t_s, t_e]$ if $F_{\mathbf{D}}(\mathcal{B}) > B_{threshold}$, where $B_{threshold}$ is a hyper-parameter as a threshold for selecting the interesting Bayesian network. Furthermore, if the temporal window $[t_s, t_e]$ is not a subset of any other interesting Bayesian network temporal window,  then  $\mathcal{B}$ is a temporal \emph{dominant interesting}  Bayesian network on that temporal window. The definition of ``dominance" provides a compact explanation.

\subsection{Discover dominant interesting Bayesian networks}
%%% Wenchong finish this part

In this section, given the independent explanation from PGM-explainer $\{\mathcal{B}_i\}_{i = t_s}^{t_e}$, and the data in each time step $\{\mathcal{D}_i\}_{i = t_s}^{t_e}$, we aim to discover interesting dominant Bayesian graph in the temporal window. We first described a brute-force approach for temporal pattern discovery. Then we described our proposed pruning approach based on the antimonocity property of dominant interesting  Bayesian 
network.  

%Temporal window pattern 
\subsubsection{Brute-force approach}
The brute-force approach has two phases, namely interesting Bayesian network discovery and dominant interesting Bayesian network discovery. In the first phase, for each candidate network, the algorithm scans every possible temporal window and computes the interest score on the window. There are $n^2$ possible temporal windows and $n$ candidate Bayesian networks. The computation cost is $O(n^3)$ for the first phase. Then in the second phase, the algorithm aims to find dominant interesting Bayesian network whose temporal window is not a subset of that of any other interesting Bayesian network. It scans every pair of interesting  Bayesian network's temporal window and eliminates the one that is not dominant. Due to the space limit, we omit the algorithm of the brute-force approach.

% \begin{algorithm}[h]
% \caption{The brute-force approach}
% \label{alg:brute-force}
% \begin{algorithmic}[1]
% \Require\quad\\
% $\bullet$ Candidate Bayesian graph $\{ \mathcal{B}_i\}_{i = t_s}^{t_e}$\\
% $\bullet$ Perturbed dataset from the model $\mathbf{D}$\\
% $\bullet$ Interest measure function $F_{\mathbf{D}}$ 
% \Ensure\quad\\
% $\bullet$ All the Interesting dominant Bayesian graph and its dominant time window
% \State{ {\bf Phase 1: Interesting Bayesian network discovery}}
% \State  {Candset $\leftarrow \emptyset$}
% \ForEach{$\mathcal{B}_i$}
%     \For{ $k \leftarrow 1$ \text{to}  $n-1$ }
%         \For{$j$ $\leftarrow k$ \text{to}  $n$ }
%             \State{  Scan the temporal window $[k, j]$ and compute \\
%             \ \ \ \ \ \ \ \ \ \ \ \ \ \  the interest measure of Bayesian network $\mathcal{B}_i$}
%             \If{$F_{\mathbf{D}}(\mathcal{B}) > B_{threshold}$}
%             \State{Add $\{\mathcal{B}_i: [k, j]\}$ to the CandSet}
%             \EndIf
%         \EndFor
%     \EndFor
% \EndFor
% \State{{\bf Phase 2: Dominant Bayesian network discovery}} 
% \For{ $i \leftarrow 1$ \text{to}  ${size(\text{CandSet})} - 1$ }
%     \For{$j$ $\leftarrow i$ \text{to}  ${size(\text{CandSet})}$ }
%         \State{Compare $\text{CandSet}[j] \text{and} \text{CandSet}[i]$, and remove the one that is a  subset of the other from the CandSet}
%     \EndFor
% \EndFor
% \textbf{return} CandSet
% \end{algorithmic}
% \end{algorithm}

\subsubsection{Proposed pruning approach} 
The brute-force approach does an exhaustive search on all temporal windows for all  Bayesian networks. There are lots of redundant computations in this step and requires the second phase to eliminate the redundant Bayesian network explanation. We propose to optimize the brute-force search by a  top-down traversal search\cite{zhou2013discovering} to leverage the temporal dominant relationship .

Specifically, given a temporal dominant Bayesian network over the temporal window $[t_1, t_2]$, we can conclude that no other temporal dominant Bayesian graph exists in the sub-temporal window $[t'_1, t'_2] \subset [t_1, t_2]$. Such property inspires a top-down traversal search. We start to calculate the interest measure of each Bayesian graph over the longest temporal window $[t_1,t_2]$. The algorithm recursively reduces the temporal window size if there is no temporal dominant Bayesian network in the temporal window $[t_1, t_2]$. Otherwise, if it finds temporal dominant Bayesian network over $[t_1, t_2]$,  the subset of the temporal window of $[t_1, t_2]$ can be pruned out for other Bayesian networks. The computation of interest measures for all Bayesian network over the subset is eliminated. 

Then we introduce the detailed implementation of the algorithm. We first construct a directed acyclic graph (DAG) to represent the 
dominant relationship between temporal windows. Each node represents one temporal window $[t_i, t_j]$ and has two child nodes that are sub-temporal windows and cover one time step less, $[t_i + 1, t_j]$ and $[t_i, t_j - 1]$. The algorithm uses a queue to do breadth first search on the temporal window. For each node, we evaluate the interest measure for all Bayesian networks over the temporal window. If the node identifies at least one temporal dominant Bayesian network, all the successors of the node will be pruned, otherwise, the algorithm continues evaluating the successors of the node. 

{\bf Computational complexity analysis: }
The top-down search approach can potentially eliminate redundant temporal search. In the best case, the algorithm identifies the longest temporal window $[t_s, t_e]$ as the temporal dominant Bayesian network. The computation for one Bayesian network is $O(n)$ for all Bayesian networks. In the worst case, no dominant Bayesian graph is identified and the algorithm needs to evaluate all $n^2$ temporal windows. The computational cost for $n$ candidate Bayesian networks is $O(n^3)$. %\mt{The brute force also mentioned $O(n^3)$?}

\begin{algorithm}[h]
\caption{The pruning approach}
\label{alg:pruning}
\begin{algorithmic}
\Require\quad\\
$\bullet$ Candidate Bayesian graph $\{ \mathcal{B}_i\}_{i = t_s}^{t_e}$\\
$\bullet$ Perturbed dataset from the model $\mathbf{D}$\\
$\bullet$ Interest measure function $F_{\mathbf{D}}$ 
\Ensure\quad\\
$\bullet$ All the Interesting dominant Bayesian graph and its dominant time window
\State  {Candset $\leftarrow \emptyset$, Create a empty queue Q}
\State{Q.enque($[t_s, t_e]$)}
\While{Q not empty}

$\mathcal{W}=[t_1, t_2] \leftarrow Q.deque()$

\ForEach{$\mathcal{B}_i$}
\State{Scan the temporal window $\mathcal{W}$ and compute \\
            \hspace{0.38in}  the interest measure of Bayesian network $\mathcal{B}$}
            \If{$F_{\mathbf{D}}(\mathcal{B}_i) > B_{threshold}$}
            \State{Add $\{\mathcal{B}_i: [t_1, t_2]\}$ to the CandSet}
            \State{Prune all subtemporal window of $\mathcal{W}$}
            
            \hspace{0.3 in} Break
            \EndIf
\If{$\mathcal{W}$ has sub-temporal window}
\State{$\mathcal{W}_1 =[t_1 + 1, t_2], \mathcal{W}_2 =[t_1, t_2 - 1] $}
\State{Q.enque($\mathcal{W}_1 $), Q.enque($\mathcal{W}_2$)}
\EndIf

    % \For{ $i \leftarrow 1$ \text{to}  $n-1$ }
    %     \For{$j$ $\leftarrow i$ \text{to}  $n$}
            
    %     \EndFor
    % \EndFor
\EndFor
\EndWhile
\textbf{return} CandSet

\end{algorithmic}
\end{algorithm}
\section{Experiment Results}
%%Wenchong writes the dataset,  , case study, 
%%%Minh writes the PGM-explainer experiment setup. 
In this section, we aim to evaluate our proposed approach on the temporal graph neural network model and present the interpretation results on the transportation domain dataset. 
\subsection{Dataset Description}
%1. Dataset: 
We use the datasets from transportation domain, which contains the traffic speed information on the road network. The dataset SZ-taxi contains the taxi trajectories of Shenzhen from Jan.1, 2015 to  Jan.30, 2015. Following the experiments setup, we select 156 major roads of Lanzhou districts as the study area. Each road segment is represented with one node.  The adjacency matrix describes the spatial relationship between roads. The feature matrix describes the speed dynamics overtime on each road. The traffic speed was aggregated every 15 minutes. The dataset was split into $60\%$ for training, $20\%$ for validation and $20\%$ for testing. We selected one  day (discretize into 96 temporal steps) to do the interpretation of the prediction on the test dataset. 

% $\bullet$ Los-loop: This dataset was collected in the highway
% of Los Angeles County in real time by loop detectors. We selected 207 sensors and its traffic speed from Mar.1
% to Mar.7, 2012. We aggregated the traffic speed every 5
% minutes

\subsection{Experiment setup}
We train T-GCN model\cite{zhao2019t} using Adam optimizer and set the learning rate to $0.001$, batch size to 64 and training epoch to 3000. The accuracy of T-GCN on the test dataset is $0.71$. Then we generate the perturbation dataset using $1000$ samples. The perturbation threshold is set to $0.01$, and perturbation probability is $0.2$. 
For dominant Bayesian network discovery, the Bayesian score threshold is $1400$ based on the validation dataset. All experiments are done on the high performance cluster using 1 NVIDIA A100 GPU with 80 GB GPU  memory. %\mt{Our definition of being interest depends on  $B_{threshold}$, thus selecting this value is quite important. Do we have any analysis on this?}

\subsection{Interpretation of results}

We present the interpretation results on some representative nodes to visualize the results. The chosen target nodes have around ten two-hop neighbors. The time series of the target node contains both congestion (low speed) and non-congestion (high speed) time slots in one day to be representative for the traffic condition.

\begin{figure}
    \centering
    \subfloat[]{\includegraphics[height=1.15in]{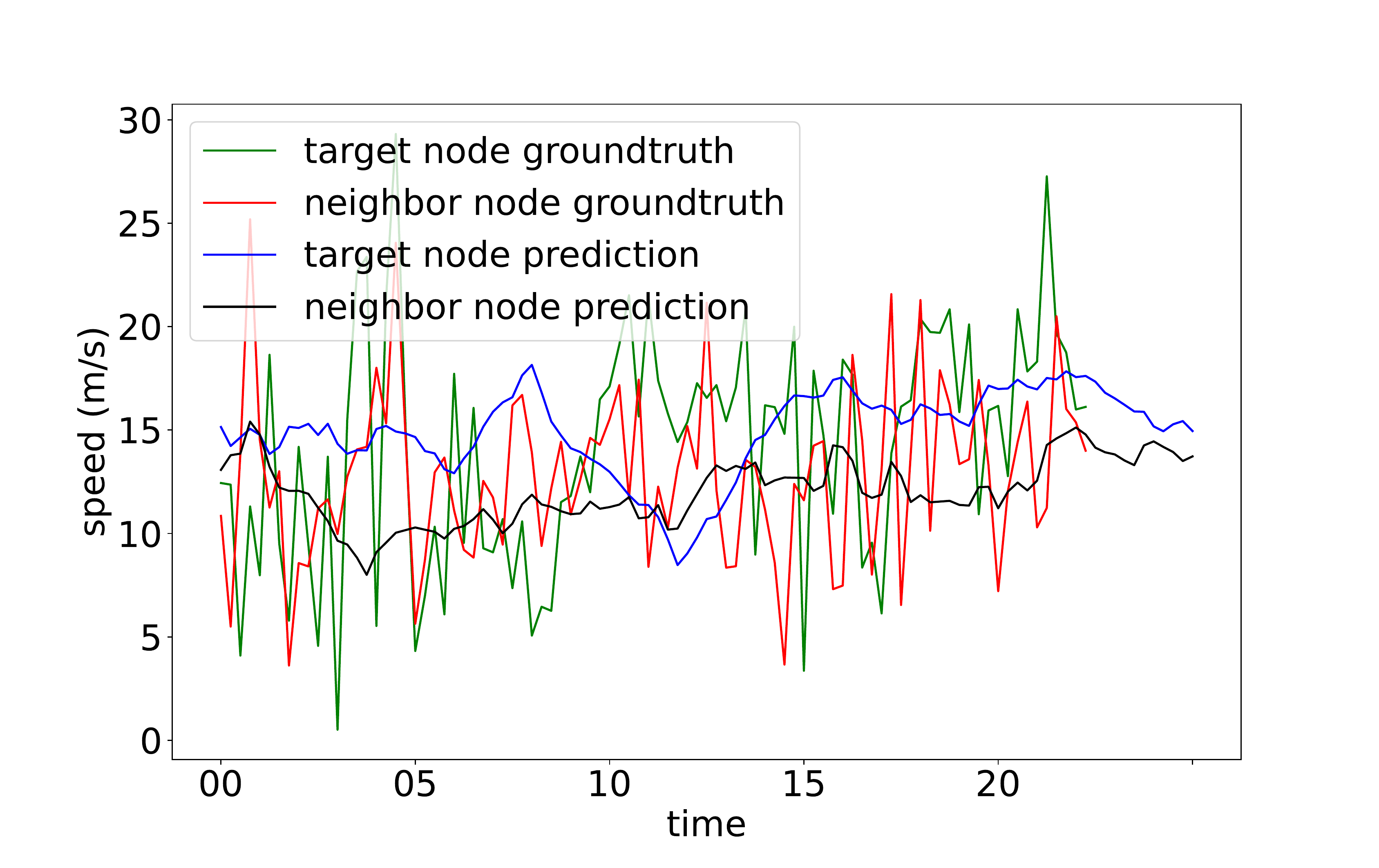}} 
    \subfloat[]{\includegraphics[height=1.15in]{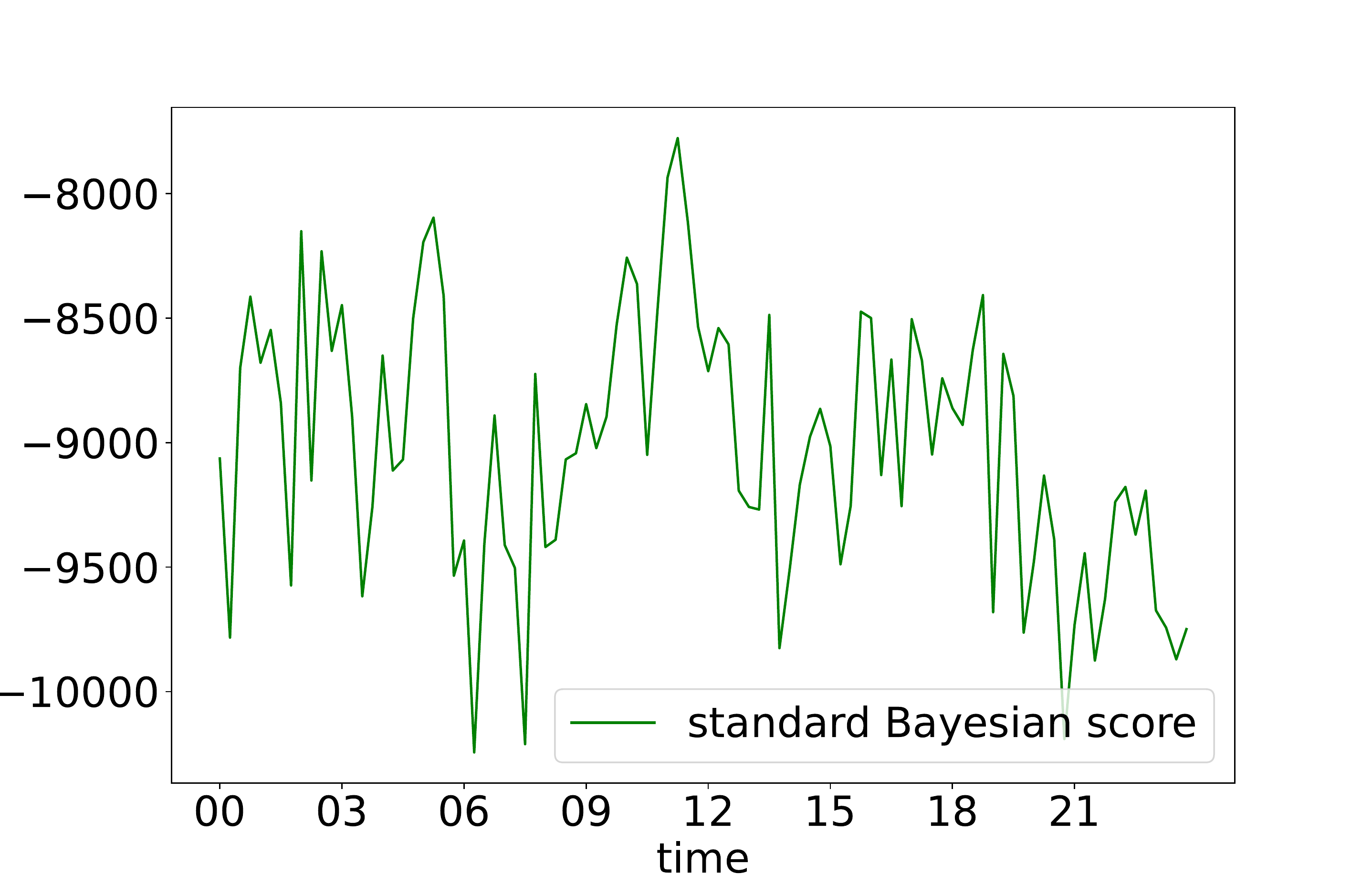}}
    \caption{Ground-truth and prediction of one target node and neighbor (a), standard Bayesian  score of the target node (b) (The x axis represents the hour in one day).} %{\zj{The legend font size is too small, units are not provided}} }
    \label{fig:node125}
\end{figure}

Figure~\ref{fig:node125}(a) shows one target node speed ground-truth and the prediction of T-GCN, as well as that of one neighbor node. From the  prediction trend, we can observe, the T-GCN model predicts worse at the peak, because the GCN model uses a smooth filter in the spatial network and capture the spatial relationship by constantly moving the filter. Before applying our method to explain the prediction of T-GCN model, we inspect the dependency dynamic of the model in each snapshot. We first construct a standard Bayesian network, which contains all neighborhood nodes within 2-hop and no edges. The Bayesian score of  standard network on each temporal snapshot data is shown in Figure~\ref{fig:node125}(b). We can observe in the time $3$ to $4$ and $11$ to $12$, the standard Bayesian score is  higher, it implies that the target node and neighbors have a higher probability to be independent of each other. On the contrary, around $20$ to $22$ the standard Bayesian score is lower, meaning there exists an interesting dependency between the target nodes and neighbors. 

\begin{figure}
    \centering
    {\includegraphics[height=1.6in]{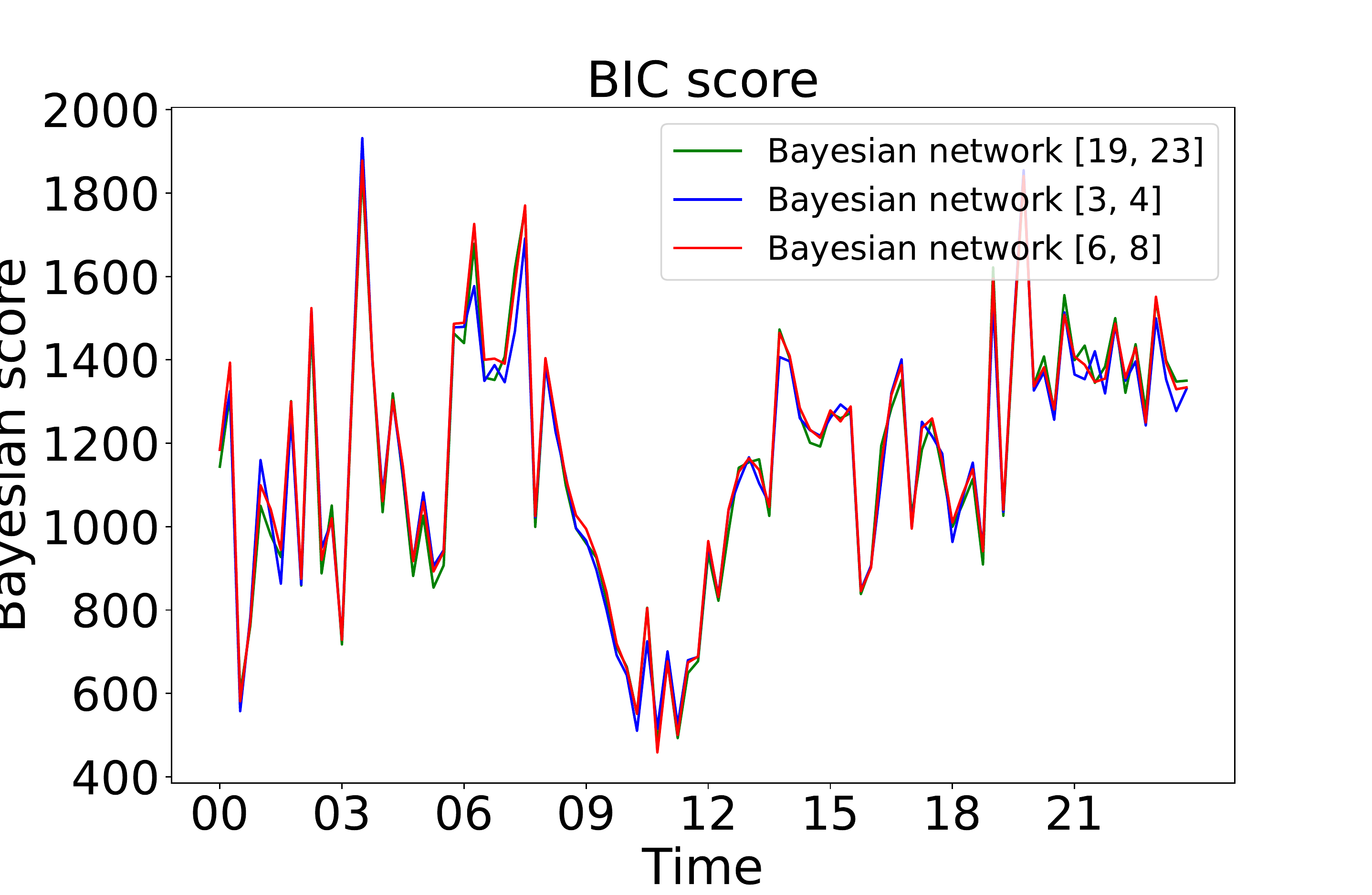}} 
    \caption{The Bayesian score evaluated at the day for three candidate Bayesian graph}
    \label{fig:node125bs}
\end{figure}

\begin{figure}
    \centering
    \subfloat[]{\includegraphics[height=0.8in]{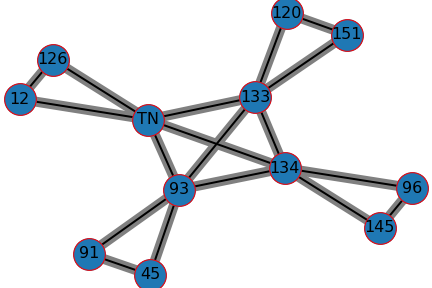}} 
    \subfloat[]{\includegraphics[height=0.8in]{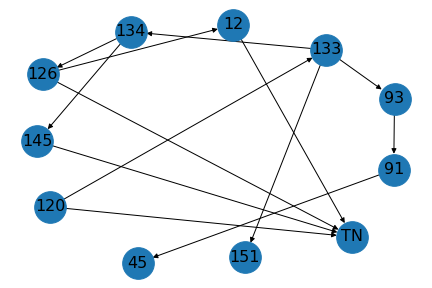}} \\
    \subfloat[]{\includegraphics[height=0.8in]{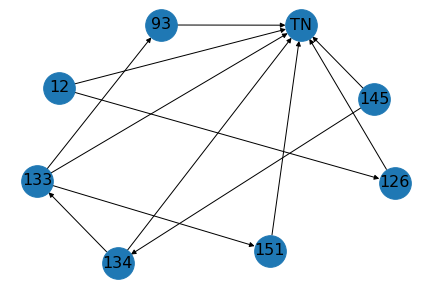}} 
    \subfloat[]{\includegraphics[height=0.8in]{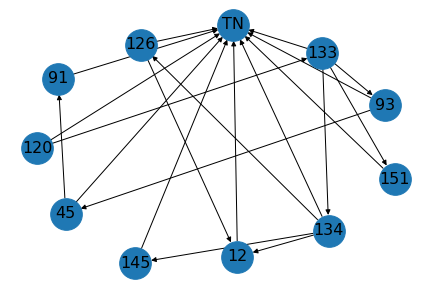}} 

    \caption{(a). Original two-hop neighbors of the target node (TN) , (b-d). Interesting dominant Bayesian discovered in temporal window: (b). [3, 4],  (c). [6, 8], (d). [19, 23]}
    \label{fig:node125explain}
\end{figure}

Secondly, we apply our approach to explain the predictions of T-GCN on each temporal snapshot during a day and obtain the Bayesian network $\{\mathcal{B}_i\}_{i = t_s}^{t_n}$. Then, we measure the fitness of the identified Bayesian network to the data in other time slots using $\textup{score}(\mathcal{B}:D_{i})$. To evaluate the relative improvement of adding the dependency edges, we use the difference of the Bayesian score between each network $\mathcal{B}_i$ and the standard  Bayesian network $\mathcal{B}_0$. Figure~\ref{fig:node125bs} shows the score results of three dominant interesting Bayesian network.  We can observe around the time window $3$ to $4$, $6$ to $8$ and $19$ to $23$, the average score is higher.  We show the explanation results of the  target node in Figure~\ref{fig:node125explain}.  Three  dominant Bayesian network were discovered in the time window [3, 4], [6, 8] and [19, 23]. We can observe that in the early morning period, the influential nodes to the target node is nodes $12, 126, 145$, while in the evening period, more dependency structure is observed. Almost all neighbor nodes within 2-hop have high influence on the target node in the evening. Based on the discovery, it is reasonable to hypothesize that the early morning period traffic in this target node is less rush than in the evening. 

\begin{figure}
    \centering
    \subfloat[]{\includegraphics[height=0.8in]{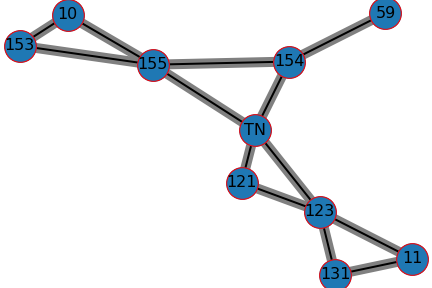}} 
    \subfloat[]{\includegraphics[height=0.8in]{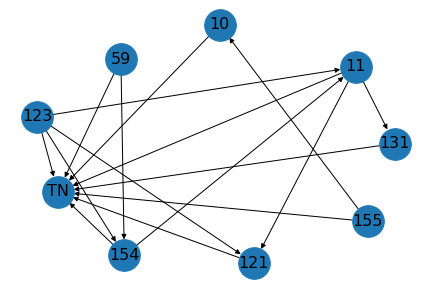}} \\
    \subfloat[]{\includegraphics[height=0.8in]{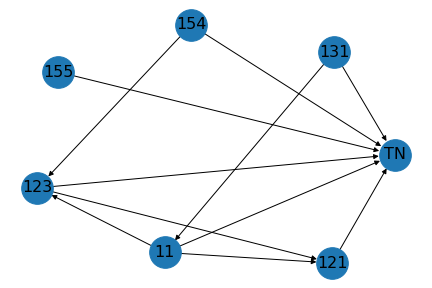}} 

    \caption{(a). Original two-hop neighbors of the target node (TN). Interesting dominant Bayesian network discovered in temporal window: (b). [8, 10],  (c). [21, 23].}
    \label{fig:node122explain}
\end{figure}

As a comparison we provide the explanation results for another target node in Figure~\ref{fig:node122explain}. We can observe for this target node, the dominant Bayesian network discovered around $[8, 10]$ (Figure~\ref{fig:node122explain} (b)) has much more dependency structure than the dominant Bayesian network in $[21, 23]$ (Figure~\ref{fig:node122explain} (c)), so we  hypothesize that for this target node,  the time window $[8, 10]$ maybe rush-hour period. 

% {\zj{We may need to have some computational experiments since we list both a brute force algorithm and a smarter algorithm}}
%4. Comparison with related work
\textit{Analysis of Bayesian score threshold}: 
The selection of interesting Bayesian network depends on the Bayesian score threshold. A higher threshold can find more critical patterns of the network dependency structure, but the related dominant windows are shorter. On the other hand, a lower threshold can find dominant Bayesian network with a longer time window but there may be redundancy. The selection of the threshold can be based on the above trade-off and domain knowledge. 

\subsection{Analysis of Computational costs}
\begin{figure}
    \centering
    {\includegraphics[height=1.5in]{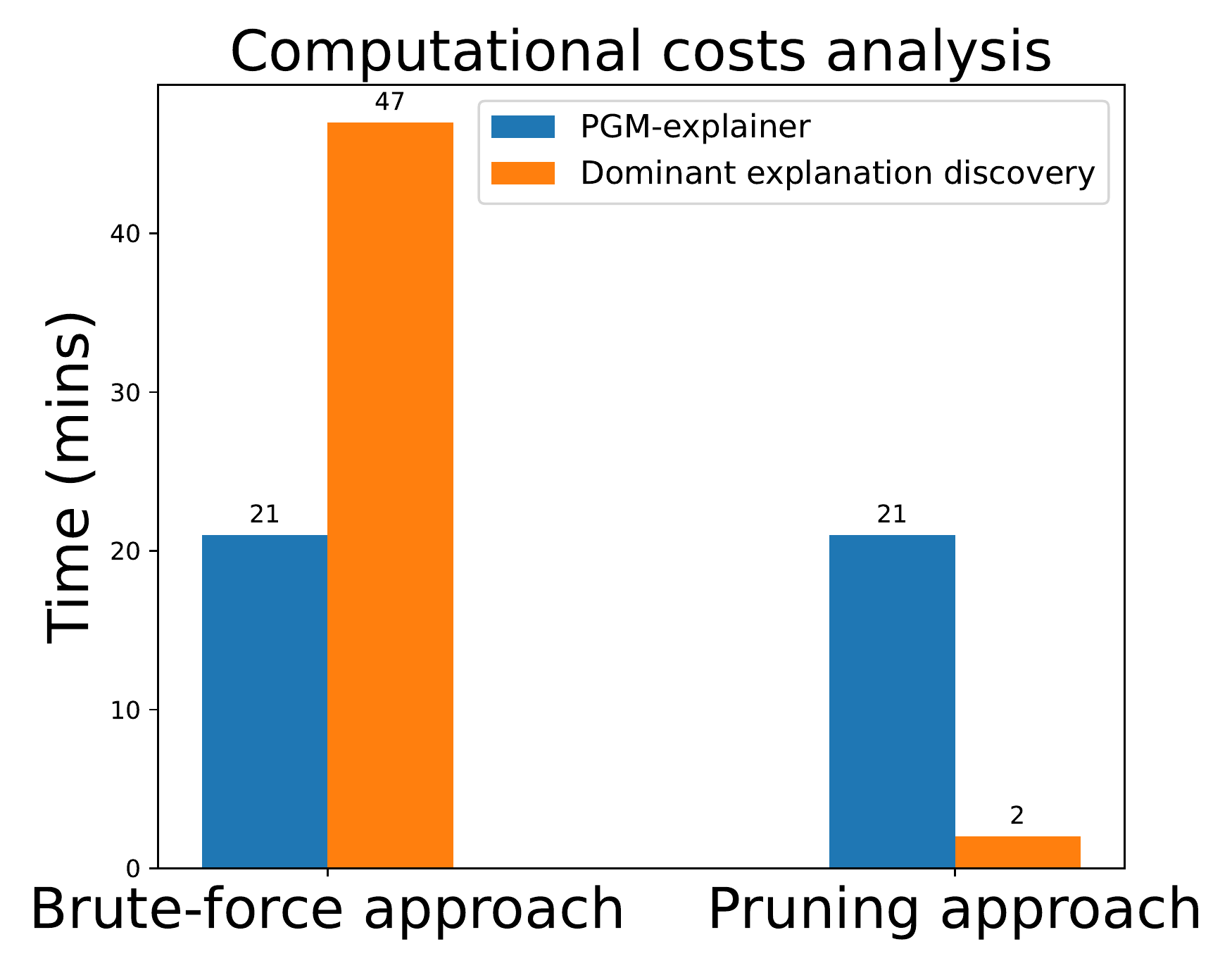}} 
    \caption{Computational cost analysis}
    \label{fig:computation}
\end{figure}

Then we analyze the computational costs for brute-force and pruning approach. As Figure~\ref{fig:computation} shows, the PGM-explainer module requires the same time for two approaches ($22$ mins). For dominant explanation discovery module, our pruning approach improve significantly than the brute-force approach.

\section{Conclusion}
In this paper, we propose a novel explainer framework for TGNN model. The framework consists of two modules that first explain each time slot prediction independently and then discovery the dominant interesting explanations in a time period. Case studies show that the proposed approach can discover the dynamic structure depenency in TGNN model.

\section*{Acknowledgement}
This material is based upon work supported by the National Science Foundation (NSF) under Grant No. FAI-1939725, SCH-2123809, IIS-2147908, IIS-2207072, and OAC-2152085.

\bibliographystyle{IEEEtran}
\bibliography{minhbibfile}

\end{document}